\documentclass[2017]{article}
\usepackage{spconf,amsmath,epsfig}
\usepackage{cite}
\usepackage{csquotes}
\usepackage{url}
\usepackage{amsmath}
\usepackage{placeins}

\usepackage{placeins}
\usepackage{epstopdf}
\usepackage{subcaption}
\usepackage{graphicx}
\usepackage{bm}
\usepackage{amssymb,booktabs,multirow}
\usepackage{algorithm,caption}
\usepackage[noend]{algorithmic} 
\usepackage{afterpage}
\newlength{\oldtextfloatsep}

\begin{document}\sloppy


\title{\enquote{RAPID} Regions-of-Interest Detection in Big Histopathological Images}
%
\name{Li Sulimowicz , Ishfaq Ahmad}
\address{Department of Computer Science and Engineering, University of Texas at Arlington, TX, USA\\
\textit{li.yin@mavs.uta.edu}, \textit{iahmad@cse.uta.edu}}

\maketitle
\begin{abstract}
The sheer volume and size of histopathological images (\textit{e.g}., $10^6$ MPixel) underscores the need for faster and more accurate Regions-of-interest (ROI) detection algorithms. In this paper, we propose such an algorithm, which has four main components that help achieve greater accuracy and faster speed:  First, while using coarse-to-fine topology preserving segmentation as the baseline, the proposed algorithm uses a superpixel regularity optimization scheme for avoiding irregular and extremely small superpixels. Second, the proposed technique employs a 
prediction strategy to focus only on important superpixels at finer image levels. Third, the algorithm reuses the information gained from the coarsest image level at other finer image levels. Both the second and the third components drastically lower the complexity. Fourth, the algorithm employs a highly effective parallelization scheme using adaptive data partitioning, which gains high speedup. Experimental results, conducted on the BSD500~\cite{datasetBSD} and 500 whole-slide histological images from the National Lung Screening Trial (NLST)\footnote{\url{https://biometry.nci.nih.gov/cdas/nlst/}} dataset, confirm that the proposed algorithm gained $13$ times speedup compared with the baseline, and around $160$ times compared with SLIC~\cite{achanta2012slic}, without losing accuracy.
\end{abstract}
\begin{keywords}
Superpixels, regions-of-interest detection, big histopathological image, parallel segmentation.
\end{keywords}
\section{Introduction}
\label{sec:intro}
Due to the explosive growth in the histopathological image data in both volume and size ( \textit{e.g.}, $10^6$ MPixel), the ROI detection problem has become a crucial ancillary pre-processing step for a wide range of computer-aided automated diagnosis applications. The impact of ROI detection is enormous because it saves pathologists from the job of verifying massive volumes of big images. ROI detection can be embedded into many medical applications, which include prostate cancer detection in digitized H\&E whole-slide images~\cite{ gorelick2013prostate, litjens2015automated}, nuclear and glandular structures automated detection and segmentation for grading of prostate and so on. 
However, the enormous timing complexity of ROI detection has baffled researchers for a long time.
\begin{figure}
        \begin{subfigure}[b]{0.33\columnwidth}
                \includegraphics[width=0.98\linewidth,height = 2cm]{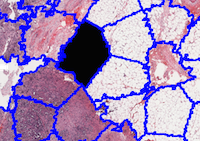}
                \caption{Proposed}
\label{fig:propose}
        \end{subfigure}%
        \begin{subfigure}[b]{0.33\columnwidth}
                \includegraphics[width=0.98\linewidth,height = 2cm]{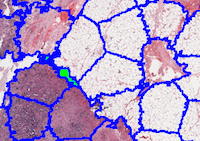}
                \caption{CTFTPS Case $1$}
                \label{fig:ct_orig}
        \end{subfigure}%
        \begin{subfigure}[b]{0.33\columnwidth}
                \includegraphics[width=.98\linewidth,height = 2cm]{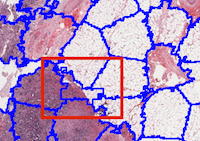}
                \caption{CTFTPS Case $2$}
                \label{fig:ct_sizelimit}
        \end{subfigure}%
        \caption{The comparison of segmentation results.}
        \label{fig:fig_com3}
\end{figure}
To overcome the computational complexity of ROI detection in big images, numerous attempts ~\cite{freund1996experiments, sertel2009computer, tabesh2005automated, lucchi2010fully,litjens2015automated,bejnordi2015multi, imageSeg2}  have been made to design more efficient algorithm. Among them, some techniques~\cite{litjens2015automated, bejnordi2015multi, imageSeg2} combined multi-resolution scheme and superpixel-driven segmentation reduced the computational burden and gained better detection accuracy compared with the techniques that employed either the single specific image resolution methods~\cite{ tabesh2005automated} or pixel-wise segmentations~\cite{lucchi2010fully}. The technique described in paper~\cite{imageSeg2} integrated the multi-scale framework into a new real-time coarse-to-fine topology preserving segmentation (CTFTPS)~\cite{ji2008automated} demonstrated its effectiveness when applied in ROI detection. CTFTPS has been demonstrated to be faster and capable of achieving better accuracy than other state-of-the-art superpixel methods~\cite{achanta2012slic,moore2008superpixel, vedaldi2008quick, levinshtein2009turbopixels}. This method~\cite{ imageSeg2} is named multi-scale CTFTPS.

However, for multi-scale CTFTPS, the addition of the multi-scale framework does not lower the computational complexity due to the following reasons: a) Neither the base CTFTPS nor the multi-scale CTFTPS distinguishes the importance of each superpixel. Instead, it keeps refining each superpixel at every stage and thus exacerbating the complexity unnecessarily; b) The multi-scale framework requires more calculations to obtain the information from each level.  Moreover, CTFTPS alone does not sustain a good balance between size regularization and boundary precision for its clustered superpixels, which may degrade the accuracy of the final classification. Without a superpixel size limit (see Fig. \ref{fig:ct_orig},), CTFTPS captures quite well the boundary, but with extremely small superpixels. However, if we limit this size to a certain range, see Fig. \ref{fig:ct_sizelimit}, CTFTPS greatly degrades the accuracy while preventing the superpixel size from reducing to be smaller than the bound. 
However, with irregular shapes and extremely small superpixels, the extracted local features  are less informative and distinctive, which lowers the classification accuracy.

In this paper, we propose a new segmentation method named RAPID (Regular and Adaptive Prediction-Induced Detection) for ROI detection, 
while using the multi-scale CTFTPS as a baseline. 
RAPID has three main features in tackling the challenges of ROI detection in big images. First, RAPID is optimized to avoid irregular and extremely small superpixels generated by CTFTPS. Simultaneously, it obtains high segmentation precision. 
Second, RAPID includes adaptive prediction-induced detection which relies on two concepts. The first is \enquote{boundary superpixel} which allows the algorithm to focus on important superpixels located at the boundary of ROI and non-ROI at finer image level. 
The second is \enquote{adaptive multi-level framework} in which the information gained from the first image level, which is a compressed version of the original image, is reused at later finer levels. The advantage is that the complexity is drastically reduced. 
As demonstrated in experiments, RAPID is $4$ times faster compared with multi-scale CTFTPS, and $46$ times compared with other state-of-the-art real-time segmentations. Third, the effective parallelization scheme using adaptive data partitioning, which leads to $4$ times additional speedup over the sequential version. 

The rest of the paper is organized as follows:  Section \ref{sec:related_work} reviews the related work. Section \ref{sec:proposed} describes the details of the proposed algorithm. Section \ref{result} presents and discusses the experimental results. The concluding remarks in Section~\ref{conclusion} provide final observations and future directions.

\section{Related work}
\label{sec:related_work}
This section gives an overview of the baseline superpixel algorithms, CTFTPS~\cite{ji2008automated} and multi-scale CTFTPS~\cite{imageSeg2}.
\subsection{Basic CTFTPS~\cite{ji2008automated}}
\label{subsec:CTFTPS}
\setlength{\intextsep}{-5pt}
\noindent\begin{align}
\label{equa:energy1}
E(\textbf{s},\bm{\mu},\textbf{c})=\sum_{p} E_{col}(s_p, c_{s_p})+\lambda_{pos}\sum_{p}E_{pos}(s_p, \mu_{s_p})\\\nonumber\noindent+E_{topo}(s)+\lambda_b \sum_{p} \sum_{q\in N_4} E_b(s_p, s_q)+\lambda_{s}E_{size}(s)
\end{align}
CTFTPS defines an energy function as above, where $s_p$ denotes superpixel label assigned to \enquote{pixel} $p$. Let $\mu_i$ be the mean  position and $c_i$ be the mean color brightness of the $i$\-th superpixel, \textbf{c} $ = (c_1, ... , c_M)$, $\bm{\mu} = (\mu_{1}, ... , \mu_{M})$, $M$ is the total number of superpixels. 
$N_4$ denotes the 4 neighbors surrounding $p$ in a $3\times 3$ block. $E_{topo}(s)$ forces superpixels to form a connected component by penalizing $+\infty$ to the function when the connectivity is not guaranteed. $E_{col}(s_p, c_{s_p}) = \|I(\tilde{p}) - c_{s_p}\|_2^2$, here $p$ is the position and $I(\tilde{p})$ is the color intensity. It is called \textbf{Appearance Coherence} and it encourages color homogeneity in each superpixel. $E_{pos} (s_p, \mu_{s_p}) = \| p - \mu_{s_p} \|_2^2$, it is called \textbf{Shape Regularity} and it imposes shape regularity to superpixels.  $E_{size}(s)$ force superpixels to be at least $1/4$ of the initial size by punishing Eq.~\ref{equa:energy1} with value $+\infty$. $E_b(s_p, s_q)$ defines  boundary pixel which has at least one neighbor pixel as defined in Eq. \ref{bound}, this term is called \textbf{Boundary Length}.
\setlength{\intextsep}{0pt}
\begin{eqnarray}
\label{bound}
E_b(s_p, s_q)=
\begin{cases}
1, &s_p \neq s_q, \cr 0, &otherwise.
\end{cases}
\end{eqnarray}
\setlength{\intextsep}{0pt}
\begin{align}
\label{operation}
\hat{s}_{p_i} &= \arg \min_{s^l_{p_i}\in N_4} E(\textbf{s}, \bm{\mu}, \textbf{c})
\end{align}
\setlength{\intextsep}{0pt}
$l$ means the block stage. CTFTPS starts by initializing superpixels with grids of different labels. At each stage, it iteratively optimizes Eq. \ref{operation} by reassigning the label of boundary blocks. It keeps investigating those boundary blocks pushed into a queue until the queue is empty. Finer stage begins on the result from previous stage by manipulating smaller blocks. 

\subsection{Multi-scale CTFTPS~\cite{imageSeg2}}
\label{subsec: multiscale_CTFTPS}

Multi-scale CTFTPS first resizes the original image into different resolutions for different levels. At each level, it performs the same refinement as CTFTPS does. The only difference is it maps the superpixel labels from the last image level to the finer one. 

\section{Details of The Proposed Algorithm}
\label{sec:proposed}


\subsection{Superpixel Regularity Optimization}
\label{subsec:merge}
For CTFTPS, there are two extreme ways to control the size of superpixels in Eq. \ref{equa:energy1}: 1) Set $\lambda_{s}=0$, there will be no restriction. 2) Set $\lambda_{s}=\infty$, the minimum size is limited to be $0.25$ of the original. The result is shown in Fig. \ref{fig:ct_orig} and  \ref{fig:ct_sizelimit}.

Clearly, case 1 has better ability to capture boundaries, but with extremely small superpixles, which in our ROI detection task can reduce the accuracy of the classifier with fragmentary information extracted from these superpixels. Case 2 avoids these small superpixels while sacrifices segmentation precision. To overcome these drawbacks, we propose a novel superpixel regularity optimization which preserves both segmentation precision and size regularity.

\textbf{Size Regularity:} The proposed works by redefining the term, $E_{size}(s)$ in Eq. \ref{equa:energy1} into a merging operation. Suppose $Size(s_{p_i})-Size(p_i)\leq l\text{ }InitSize(s_{p_i})$ superpixel ${s_{p_m}}$, $l$ is the lower bound of the size threshold,  \textit{e.g.}, we can set $l=0.25$. We search another superpixel $s_{p_n}$ which satisfies three conditions, 1) $s_{p_n}$  is a neighbor of ${s_{p_m}}$ to make sure the connectivity; 2) It has the least gross energy after merging (Eq.~\ref{eqa:merge}) to ensure the merged two superpixels are similar; 3) and  after merging, the size of  $s_{p_n}$ will not exceed the upper bound to avoid the potential problem of getting very big superpixels (Eq.~\ref{condtion}). 
\setlength{\intextsep}{0pt}
\begin{align}
\label{eqa:merge}
&\hat{s_{p_i}} =\arg\min\limits_{s_{p_i}\in N_{s}}E(\textbf{s},\textbf{c},\bm{\mu})
\\ \label{condtion}&\textit{s.t. } Size(\hat{s_{p_i}})+Size(p_i)\leq u\text{ }InitSize(\hat{s_{p_i}})
\end{align}
\setlength{\intextsep}{0pt}
Here, $N_{s}$ denotes the set of contiguous superpixels of $s_{p_i}$, and $u$ is the upper bound parameter. After merging, the algorithm can keep refinement on all the superpixels till convergence. Thus, this merging optimization will not degrade the algorithm's capability to snap the boundaries.

Instead of computing the absolute energy of before and after the merging, in practical, we compute the difference of energy between after merging and before merging since it is only related with the two superpixels that are merged, which could significantly degrade the computational complexity.

\subsection{Adaptive Prediction-Induced Detection}
\label{subsec:boundary}
\begin{algorithm}[H]
\small
\caption*{RAPID} \label{alg:RAPID}
\begin{algorithmic}[1]
\STATE  Initialize superpixels to be regular grids and assign labels;
\FOR{ $l=1$ to levelMax}
\IF{$l==2$}
\STATE Extract features and get predict vector $P$;
\ENDIF
\FOR{$s=1$ to stageMax}
\STATE Initialize each block with smaller regular grid;
\STATE Map superpixels' labels and store in $L$ matrix;
\IF{$s==1$ \&\& {$l==1$}} 
\STATE Compute initial $\hat{\mu_i}^l$ and  ${\hat{c_i}}^l$ for each superpixel and store in $S$ vector;
\ELSE
\STATE Recompute $\hat{\mu_i}^l$ according to Eq. \ref{eq:mu};
\ENDIF
\STATE Use rule from Eq. \ref{new_bound} to generate boundary queue $Q$;
\WHILE{$Q$ is not empty}
\STATE Pop out boundary block ${p_i}^{(l,s)}$ from the queue;
\IF{connectivity of ${p_i}^{(l,s)}$ is valid}
\STATE Compute mean brightness ${c_i}^{(l,s)}$ and mean position $\mu{_i}^{(l,s)}$ for block ${p_i}^{(l,s)}$ and its neighbors;
\IF{The size of ${s_{{p_i}}^{(l,s)}}$ after removing ${p_i}^{(l,s)}$ is smaller than the lower bound}
\STATE Use optimization from Eq. \ref{eqa:merge} and Eq. \ref{condtion};
\ELSE 
\STATE Use original operation from Eq. \ref{operation};
\ENDIF
\ENDIF
\IF{${p_i}^{(l,s)}$ is updated}
\STATE Update $\hat{\mu}^{(l,s)}$ and $\hat{c}^{(l,s)}$;
\ENDIF
\ENDWHILE
\ENDFOR
\ENDFOR
\end{algorithmic}
\end{algorithm}
When CTFTPS is applied in ROI detection task, there exists a giant computational redundancy because only those superpixels located at the boundaries between ROI and non-ROI matter. However, CTFTPS cannot recognize them and keeps refining to find the precise boundaries between all the superpixels. For multi-scale CTFTPS, it does use multi-scale framework but adds more computational complexity by calculating the information of each superpixel at each image level. RAPID relies on two concepts to overcome these drawbacks, \enquote{boundary superpixel} and \enquote{adaptive multi-level framework}.

RAPID alleviates the first drawback by localizing those \enquote{boundary superpixels} at a coarse image level. The algorithm does so by utilizing the segmentation result from an earlier coarse stage with a pre-trained classification model to generate a prediction map indicating the type of superpixel. The details of this first method are given as follows: 

For ROI detection, the hypothesis representative $h_{\theta}(x)$ is defined in Eq. \ref{predict}, where the input $x$ is a vector composed of features extracted from superpixels, $y$ is the output of the classifier (denoted as $h_{\theta}(x)$). The value of $y$ equals to $1$ indicates that it is classified as ROI, and $-1$ as non-ROI. Here, we introduce a new concept, \enquote{boundary superpixel}. As shown in Eq. \ref{new_bound}, when $y_{s_p}\neq y_{s_q}$, we can call such a superpixel a \enquote{boundary superpixel}.
\setlength{\intextsep}{0pt}
\begin{flalign}
\label{predict}
\noindent y&=
\begin{cases}
1,&h_{\theta}(x)\geq 0,\cr -1,&otherwise.
\end{cases}
\end{flalign}
  \vspace{-0.5cm}
\begin{eqnarray}
\label{new_bound}
\noindent E_b(s_p, s_q)&= 
\begin{cases}
1,&s_p \neq s_q, y_{s_p}\neq y_{s_q},
\cr0, &otherwise.
\end{cases}
\end{eqnarray}
\setlength{\intextsep}{0pt}
  \begin{figure}
        \begin{subfigure}[b]{0.49\columnwidth}
                \includegraphics[width=0.97\linewidth, height=3.1cm]{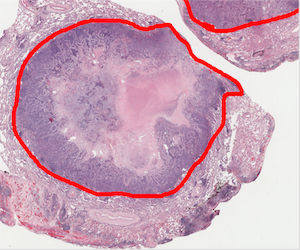}
                \caption{Ground Truth}
                \label{fig:fig_s1}
        \end{subfigure}%
        \begin{subfigure}[b]{0.49\columnwidth}
                \includegraphics[width=0.97\linewidth,height=3.1cm]{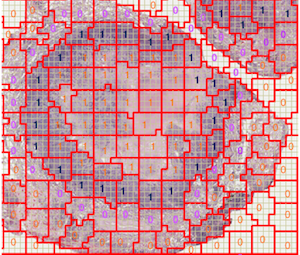}
                \caption{Prediction Map}
                \label{fig:fig_s2}
        \end{subfigure}%
        \caption{Prediction map is the prediction result of the classifier in an intermediate stage. $0$ denotes non-ROI, $1$ denotes ROI. And we used purple-colored $0$ and blue-colored $1$ to represent those \enquote{boundary superpixels}.}
        \label{fig:fig_speed}
\end{figure}
\setlength{\intextsep}{0pt}


The speed up method utilizes Eq.~\ref{new_bound} to strengthen the original restriction defined in Eq.~\ref{bound}. Only by putting those blocks located in \enquote{boundary superpixel} in the queue at the later stages, RAPID decreases the length of the boundary queue while at the same time preserves the detection accuracy.  The effect is shown in Fig. \ref{fig:fig_speed}. The ground truth of the ROI is shown in Fig. \ref{fig:fig_s1}. Fig. \ref{fig:fig_s2} shows the prediction map in an intermediate stage of RAPID. 
RAPID only needs to refine the \enquote{boundary superpixels} marked by purple-colored $0$ and blue-colored $1$ instead of refining all of them as the CTFTPS did.

Moreover, RAPID eliminates the second drawback by only finding the information at the first image level. At the left over image levels, it reuses the information of the upper level with simple deduction. The details are explained below:

\textbf{Adaptive Multi-level Framework: }RAPID eliminates the cost of gaining the information of each image at different level by finding only the average information of each superpixel at the first image level. At the next image level, RAPID directly uses the average brightness information of the upper level. For the position information, it only needs to perform the following operation:
\setlength{\intextsep}{0pt}
\begin{align}
\label{eq:mu}
\mu_{i+1} = C\times \mu_i - \frac{C-1}{2}
\end{align}

Here, $C$ is the parameter of the compression ratio, $i$ refers to the image level. 
This makes the cost of obtaining average information for superpixels from original image only ${1}/{C^L}$ of the original cost. $L$ is the total levels.

The pseudocode for RAPID is shown above. 

\subsection{Parallel RAPID}
\label{subsec:parallel}

Recall that the four main tasks in RAPID are:  1) Mean color and position computation for each superpixel on vector $S$; 2) Label mapping from an upper stage on matrix $L$;  3) Going through matrix $L$ to find the boundary queue $Q$ and 4) Relabeling in $Q$. For efficient parallelization, we aim to execute all four tasks in parallel by exploiting data parallelism through partitioning the image into tiles. Task 2 can be trivially parallelized while task 1 involves updating a vector which requires locking mechanism on a shared memory multi-core machine. Luckily, benefit from the adaptive multi-level framework, only the first image level includes this task. Relabeling involves tackling the contention on shared variables of superpixels when data is partitioned across multiple parallel threads. We utilized the OpenMP~\cite{dagum1998openmp} to implement the Parallel RAPID on a shared-memory machine.

\textbf{Parallel Split:} is used to decrease the conflict in task 4. It works by evenly split the number of rows in whole image as shown in Eq.~\ref{parallel_spit}. $i$ is the current thread number. Every data tile is processed by a single thread, and finds its own boundary queue, $Q_i$ and performs relabeling. The pseudocode of parallelized RAPID is given as follows.
\setlength{\intextsep}{0pt}
\begin{align}
\label{parallel_spit}
startRow &= totalRow * i / totalThreads\\\nonumber
endRow &= totalRow * (i + 1) / totalThreads
 \end{align}
 \setlength{\intextsep}{-15pt}
 \begin{algorithm}[H]
 \small
\ \nonumber
\caption*{Parallel RAPID} \label{alg:alg_parallel}
\begin{algorithmic}[1]
\FOR { $l=1$ to levelMax}
\IF{$l==2$}
\STATE Extract features and get predict vector $P$;
\ENDIF
\FOR{$s=1$ to stageMax}
\STATE Trivially parallelize label mapping and store in $L$;
\IF{$s==1$}
\STATE Parallelize computing initial $\hat{\mu_i}^l$ and  ${\hat{c_i}}^l$ for each superpixel $i$ and store in $S$ vector (write \& read lock protection on $S$);
\ENDIF
\STATE Start $N$ threads;
\FOR { $i = 0$ to $N-1$ }
\STATE  Do parallel split as Eq. \ref{parallel_spit} and Finding the queue $Q_i$;
 \STATE Dealing $Q_i$ independently in each thread or tile with Locking mechanism for $S$ vector;
\ENDFOR
\ENDFOR
\ENDFOR
\end{algorithmic}
\end{algorithm}

\subsection{Computational Complexity Analysis}
\label{subsec:complexity}
For CTFTPS and multi-scale CTFTPS, the computational complexity can be summarized with $O(\sum_l(\sum_i{{Q_{l}}^{i}}+\sum_i{{L_{l}}^{i}}+{S_l}))$. $i$, $l$ denotes different block stage and image level, $Q$ is the number of blocks in the queue, $L$ is the label mapping from the upper stage and $S$ means the initialization cost of computing each superpixel's mean color and position. For CTFTPS, $l$ equals to one. 
For RAPID,
 the computational complexity of RAPID is $O(\sum_l(\sum_i{\sqrt{{Q_{l}}^{i}}}+\sum_i{{L_{l}}^{i}})+{S_1})$. 

\section{Experiments}
\label{result}

\subsection{Experimental Setup}
\label{setup}
We used two datasets: 1) The Berkeley segmentation data set (BSD500)~\cite{datasetBSD},  a standard dataset to evaluate segmentation algorithms, given that it provides multiple well-labeled ground truth contours and segmentations. 2) The NLST dataset of 500 whole slide histopathological images with ROI ground truth. It is divided into two sets, training dataset with $400$ images, and testing dataset which includes the left $100$ images.

To perform ROI detection, after segmentation, we used LIBSVM package\footnote{\url{https://www.csie.ntu.edu.tw/~cjlin/libsvm/}} to classify the superpixles into two categories, ROI and non-ROI. The feature vector is denoted as $F\in {\Re}^4$, which includes absolute and comparative RGB brightness, as well as extra-region and intra-region dissimilarity~\cite{ren2003learning}. $1/2$ positive and $1/2$ negative randomized samples are used in case of overfitting in the classification.

The hardware platform is the Intel(R) Xeon(R) CPU E$5$-$2620$ v$2$ @ $2.10$GHz with $12$ cores, each core with $2$ threads. 
\begin{figure*}[!htb]
	\centering 
	\noindent \makebox[\textwidth][c]{\includegraphics[width=1.0\textwidth, height=4cm]{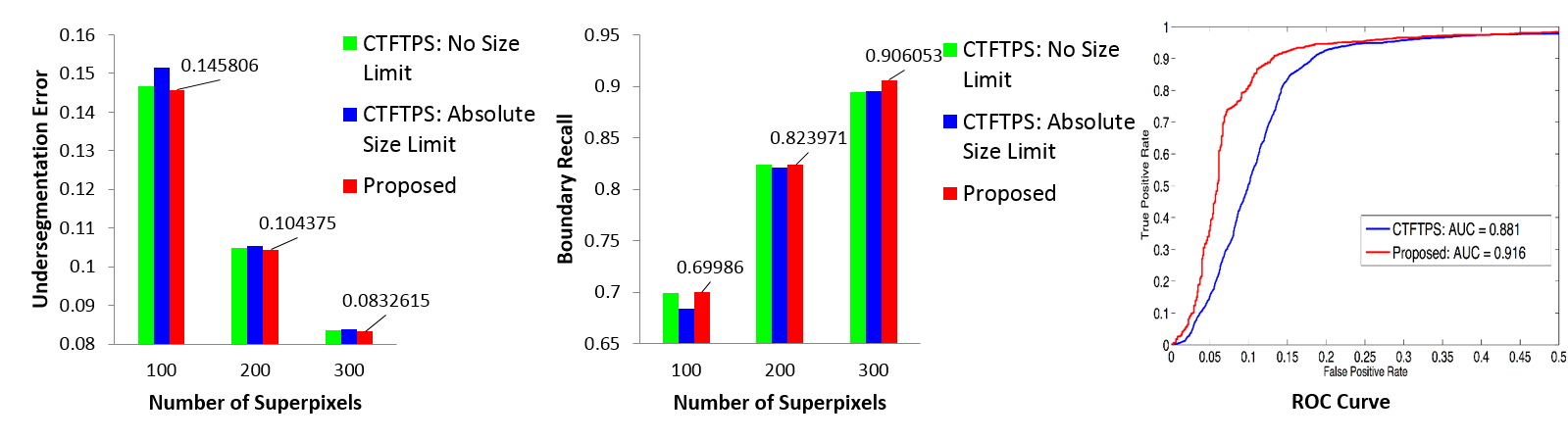}}
\caption{Comparison between CTFTPS with and without superpixel regularity optimization. }
	\label{figure:merge}
\end{figure*} 
\begin{figure*}[!htb]
\centering
        \begin{subfigure}[b]{0.33\textwidth}
                \includegraphics[width=0.97\linewidth, height=3.4cm]{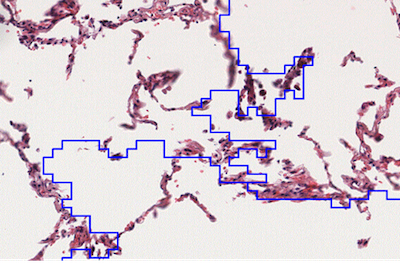}
                \caption{Coarse segment in non-ROI}
                \label{figure:cell_bound1}
        \end{subfigure}%
        \begin{subfigure}[b]{0.33\textwidth}
                \includegraphics[width=0.97\linewidth, height=3.4cm]{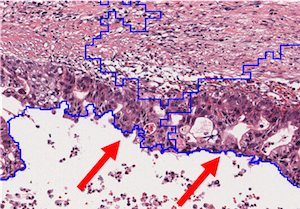}
                \caption{Fine segment at the boundary}
                \label{figure:cell_bound2}
        \end{subfigure}%
        \begin{subfigure}[b]{0.33\textwidth}
                \includegraphics[width=.97\linewidth,height=3.4cm]{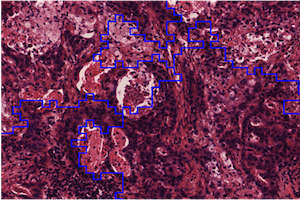}
                \caption{Coarse segment in ROI}
                \label{figure:cell_bound3}
        \end{subfigure}%
       \caption{Segmentation boundaries in different area of RAPID.}
        \label{figure:cell_bound}
\end{figure*}

\subsection{Evaluation of Superpixel Regularity Optimization}
\label{subsec:sro}
\textbf{On BSD500: } 
We applied superpixel regularity optimization on CTFTPS to segment 200 images from BSD500 into varied number of superpixels and used standard metrics~\cite{Peer}, under-segmenation error (UE) and boundary recall (BR), to evaluate the performance of superpixel regularity optimization. UE measures the percentage of pixels that leak from ground truth boundaries. BR measures the ratio of ground truth boundaries correctly covered by the superpixels. Higher score of BR and lower score of UE is expected here.

The result from Fig. \ref{figure:merge} shows that CTFTPS with our proposed method gained lower UE score and higher BR score compared with the baseline. This result demonstrated that as a segmentation method, the proposed improves the accuracy upon the original CTFTPS to snap boundaries. 

\textbf{On NLST: } We got a resized NLST dataset (compressed into size around $1000\times1000$) and used CTFTPS combined with superpixel regularity optimization to segment the $400$ images from the training set. Further, we extracted features from these superpixels and trained a  SVM classifier. The proposed generated superpixels with size in the range of $(0.24, 1.50)$. We compared the detection accuracy from the test set between the unoptimized and optimized CTFTPS in terms of the ROC curve. As shown in Fig. \ref{figure:merge}, the ptimization method gains more desirable ROC curve. The visual result shown in Fig. \ref{fig:propose} indicates that the proposed method effectively improve the classification accuracy with more and regular superpixels.

\subsection{Evaluation of RAPID}
\label{subsec:su}


We prepared two different sized NLST datasets, one dataset wherein each image is resized to the one-fifth of the original image size, the other is with the original size. We totally performed three sets of experiments. We conducted the first set of experiments on the resized dataset, and the second and third set of experiments on original dataset. In the first two experiments, the final block size was $1\times1$. The third set of experiment was also on the original dataset but with final block size as $4\times4$. We ran optimized multi-scale CTFTPS to obtain a pre-trained classifier which is used for testing RAPID.  

\textbf{Speed:} We compared RAPID with SLIC, CTFTPS, and multi-scale CTFTPS. The average time cost is shown in Table~\ref{table:table2}, indicating that our method gained $46$ times acceleration compared with SLIC, and $4$ times compared with multi-scale CTFTPS. The adaptive prediction-induced detection method attained a processing rate of $40$ MPixels/s, if minor loss is allowed ( $4\times4$ block size ), the processing speed could go up to $100$ MPixel/s. On the resized dataset, the speed is $18$ MPixel/s, which implies that RAPID is more advantageous for big images.

\textbf{Accuracy:} We ran CTFTPS, multi-scale CTFTPS, optimized multi-scale CTFTPS, on the training set of the original dataset, and then trained them through the same SVM classifier. RAPID as well as the other  algorithms were tested on the same testing set. RAPID used the pre-trained hypothesis model gained from the optimized multi-scale CTFTPS both in segmentation and testing stage.  The other algorithms used the classifier model trained by their own segmentation result. The classification accuracy on the testing set is shown in Table \ref{tab:precision}, indicating that RAPID delivers better accuracy than CTFTPS and about the same accuracy as the multi-scale CTFTPS. The results also indicate that CTFTPS has the lowest precision. CTFTPS does gain a higher precision for smaller images, but for big images it needs more stages to ensure the refinement process which makes it difficult to set up the parameters in the function. The results also shows that RAPID detects the ROI quickly with only minor precision loss.  Fig. \ref{figure:cell_bound} demonstrates the superpixel boundary in different areas, ROI, non-ROI, and the boundary. Coarse segmentation is observed in both non-ROI and ROI, which is reasonable because the segmentation in the non-boundary area ceases earlier than at a boundary area. Fig. \ref{figure:cell_bound2} shows that a very fine segmentation is obtained for \enquote{boundary superpixles}, which positively impacts the final segmentation precision. 
\vspace{5pt}
\setlength{\textfloatsep}{25pt}
\begin{table}[!ht]
\small
\centering
\noindent \captionof{table}{Comparison of the average time cost.}
\noindent \begin{tabular}
{p{0.23\columnwidth}| p{0.08\columnwidth}| p{0.09\columnwidth}| p{0.12\columnwidth}| p{0.12\columnwidth} |p{0.09\columnwidth} } 
\hline
{Average Size} &Grain& {\text{SLIC} }& {CTFTPS} &{Multi-CTFTPS}& \textbf{RAPID} \\ [0.5ex] 
\hline\hline
$4712\times5867$&$1\times1$ & $12.90s$  &$3.29s$&$ 3.54s$& $\mathbf{1.56s}$  \\ [1ex] 
 \hline
\multirow{2}{*} {$23561\times29335$}   &$4\times4$&N/A& $27.26s$&$28.98s$&$\mathbf{6.96s}$\\
&$1\times1 $& $809s $& $79.46s$&$66.46s$&$\mathbf{17.54s}$ \\ [1ex] 
 \hline\hline
\end{tabular}
 \label{table:table2}
\noindent\captionof{table}{ Comparison of the detection accuracy.}
  \noindent \begin{tabular}{p{0.15\columnwidth}|p{0.12\columnwidth}| p{0.15\columnwidth}| p{0.26\columnwidth}|p{0.1\columnwidth}} 
    \hline
  	& CTFTPS&Multi-CTFTPS  & Optimized Multi-CTFTPS&\textbf{RAPID}  \\ 
	\hline\hline
Precision &0.794& 0.818&$\mathbf{0.838 }$&$\mathbf{0.816}$ \\ \hline
F1 Score &0.877&  0.901&$\mathbf{0.912}$ &$\mathbf{0.898}$ \\
 \hline\hline
  \end{tabular}
   \label{tab:precision}
\afterpage{\global\setlength{\textfloatsep}{\oldtextfloatsep}}
\end{table}
\setlength{\textfloatsep}{-5pt}
\subsection{Evaluation of Parallel RAPID}
\label{subsec:e_parallel}
\begin{figure}
        \begin{subfigure}[b]{0.48\columnwidth}
                \includegraphics[width=1\linewidth]{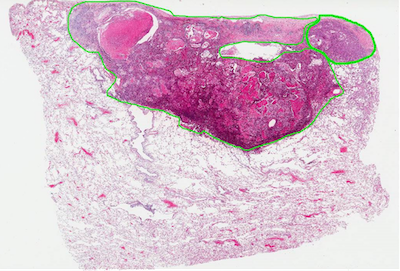}
                \caption{Ground Truth}
                \label{figure:detection1}
        \end{subfigure}%
        \begin{subfigure}[b]{0.48\columnwidth}
                \includegraphics[width=1\linewidth]{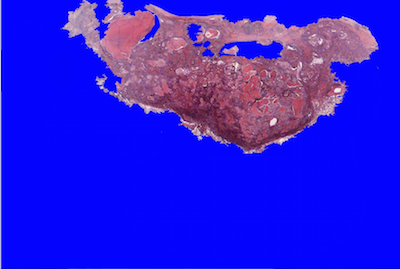}
                \caption{Detected ROI}
                \label{figure:detection2}
        \end{subfigure}%
       \caption{ROI detection result of RAPID.}
        \label{figure:detection}
\end{figure}

%
We used the same datasets and experiments from Section \ref{subsec:su}. Table \ref{table:table3} shows the average running time with different number of threads.  Note that the maximum processing speed is up to $449$ MPixel/s with grain $4$, and $133$ MPixel/s with grain $1$, which is nearly $160$ times faster than the serial SLIC, whose processing ability is $0.85$ MPixel/s, nearly $13$ times faster compared with multi-scale CTFTPS. The final visual result is shown in Fig. \ref{figure:detection}. In conclusion, Parallel RAPID improved the speed without sacrificing the accuracy of ROI detection. 

\setlength{\intextsep}{0pt}

\setlength{\intextsep}{0pt}
\begin{table}[!ht]
\small
\begin{center}
\centering
 \captionof{table}{Average time cost of Parallel RAPID with different number of threads.}
\noindent \begin{tabular}{p{0.22\columnwidth}| p{0.1\columnwidth} |p{0.08\columnwidth}| p{0.08\columnwidth}| p{0.08\columnwidth}| p{0.08\columnwidth} |p{0.08\columnwidth} } 
\hline
{Average Size} &Grain&$2$T& {\text{$4$T} }& {$8$T} &{$14$T}&$24$T\\ [0.5ex] 
\hline\hline
$4712\times5867$ &$1\times1$ &$1.09s$ & $0.63s$  &$0.49s$&$ 0.42s$& $0.34s$  \\ [1ex] 
 \hline
\multirow{2}{*} 
{$23561\times29335$}  &$4\times4 $& $4.39s $& $3.79s$&$1.95s$& $1.69s$ &$1.54s$\\[1ex]

 &$1\times1$ &$12.13s$&$7.83s$& $6.29s$&$5.83s$&$5.20s$\\[1ex]
 \hline\hline
\end{tabular}
 \label{table:table3}
 \end{center}
\end{table}



\section{Conclusions}
\label{conclusion}
As shown by the experiments, the proposed algorithm is \enquote{rapid} and accurate. It preserves both the segmentation precision and size regularity with superpixel regularity optimization. Moreover, RAPID has significantly lower computational complexity with only a negligible loss of the ROI detection accuracy, compared to the refered algorithms. Furthermore, Parallel RAPID obtained consistent speedup. Except that the Prediction-Induced method which needs users to train task related classifier, the other three main methods can be directly used on any type of images, which make the parallel RAPID a universal and applicable segmentation algorithm for big images. Our future work is exploring superpixel grouping, which could further speedup the whole segmentation process by using smaller number of superpixels. We are also examining how to enhance the scalability of the Parallel RAPID.

\bibliographystyle{IEEEbib}
\bibliography{icme2017template}

\end{document}